\documentclass[twoside,11pt]{article}
\usepackage{jair, theapa, rawfonts}

\usepackage{graphicx}%
\usepackage{multirow}%
\usepackage{amsmath,amssymb,amsfonts}%
\usepackage{amsthm}%
\usepackage{mathrsfs}%
\usepackage[title]{appendix}%
\usepackage{xcolor}%
\usepackage{textcomp}%
\usepackage{manyfoot}%
\usepackage{booktabs}%
\usepackage{algorithm}%
\usepackage{algorithmicx}
\usepackage{algpseudocode}%
\usepackage{listings}%
\usepackage{theapa}%

\usepackage{booktabs}
\usepackage{multirow}

\jairheading{1}{2025}{-}{12/25}{-}
\ShortHeadings{Efficient Continual Learning: A Low-Rank Adaptation Approach}
{Carrión, \& Casacuberta}
\firstpageno{1}

\begin{document}

\title{Efficient Continual Learning in Neural Machine Translation: A Low-Rank Adaptation Approach}

\author{\name Salvador Carrión \email salcarpo@prhlt.upv.es \\
       \addr Pattern Recognition and Human Language Technology,
       Universitat Politècnica de València\\
       \AND
       \name Francisco Casacuberta \email fcn@prhlt.upv.es \\
       \addr Valencian Graduate School and Research Network of Artificial Intelligence,
       Universitat Politècnica de València\\
       }


\maketitle

\begin{abstract}
Continual learning in Neural Machine Translation (NMT) faces the dual challenges of catastrophic forgetting and the high computational cost of retraining. This study establishes Low-Rank Adaptation (LoRA) as a parameter-efficient framework to address these challenges in dedicated NMT architectures. We first demonstrate that LoRA-based fine-tuning adapts NMT models to new languages and domains with performance on par with full-parameter techniques, while utilizing only a fraction of the parameter space. Second, we propose an interactive adaptation method using a calibrated linear combination of LoRA modules. This approach functions as a gate-free mixture of experts, enabling real-time, user-controllable adjustments to domain and style without retraining. Finally, to mitigate catastrophic forgetting, we introduce a novel gradient-based regularization strategy specifically designed for low-rank decomposition matrices. Unlike methods that regularize the full parameter set, our approach weights the penalty on the low-rank updates using historical gradient information. Experimental results indicate that this strategy efficiently preserves prior domain knowledge while facilitating the acquisition of new tasks, offering a scalable paradigm for interactive and continual NMT.
\end{abstract}

\section{Introduction}\label{sec-intro}

Neural Machine Translation (NMT) has made significant progress in recent years, achieving near-human accuracy for high-resource languages~\shortcite{seq2seq,gnmt,transformer}. However, one of the main challenges in the field is the continual learning problem, which states that a model should be able to learn from a stream of continuously incoming data while retaining the previously acquired relevant knowledge~\shortcite{thompson-cf,mccloskey-cf}. This challenge is particularly interesting in the Machine Translation domain, where the linguistic landscape is constantly evolving with new languages, terminologies, and stylistic nuances~\shortcite{cl-nmt}.

The continual learning problem in NMT is not just an academic concern; it has significant real-world applications to maintain the relevance of existing machine learning models in dynamic and ever-evolving environments, such as post-editing and interactive translation environments, where knowledge after corrections must be quickly incorporated into these systems~\shortcite{survey-cl}. As the speed of our world increases and our communications become more globalized, we expect NMT models to be able to adapt to new dialects, jargon, and evolving languages in a continuous and seamless manner. The ability to do so efficiently and accurately is crucial for businesses, governments, and individuals who rely on accurate translations. 

The difficulty in tackling the continual learning problem in NMT comes from the tendency of neural networks to forget previously learned relevant information upon learning a new one, a phenomenon known as catastrophic forgetting~\shortcite{thompson-cf,mccloskey-cf}. This is often a result of the stability-plasticity dilemma, where increasing a model's plasticity to learn new tasks inevitably compromises its stability on previous ones~\shortcite{cf-old-massive}. Naive approaches to continual learning, such as periodic retraining on a mix of old and new data, are not feasible due to the expensive computational costs associated, and the impracticality of storing all the previous data, which is not always available~\shortcite{cf-generative-replay,few-shot-reg}. Furthermore, methods that attempt to isolate new knowledge in separate full-network parameters can lead to an inefficient allocation of resources and bloated model architectures~\shortcite{cf-da-progressive-networks}.

Past solutions have often focused on either expanding the model capacity to accommodate new information or on regularizing the learning process to mitigate the forgetting~\shortcite{cf-survey}. However, these solutions typically require a significant increase in the number of parameters, computational resources, and energy consumption. While methods like Elastic Weight Consolidation (EWC)~\shortcite{ewc} have been successful in general neural networks, applying them efficiently to the dedicated encoder-decoder architectures of NMT without incurring massive storage overheads remains a challenge. Additionally, they often involve complex mechanisms to balance the retention of old knowledge and the acquisition of new information, which can be difficult to calibrate and may still result in performance degradation on previously learned tasks.

In this work, we propose a unified solution that specifically targets the dedicated NMT architecture (non-LLM) using Low-Rank Adaptation (LoRA). The primary objectives of this paper are twofold: first, to rigorously evaluate the efficacy and boundaries of LoRA for multi-domain and multi-lingual NMT adaptation; and second, to propose a method to mitigate catastrophic forgetting within this low-rank framework. While LoRA has been explored in general continual learning (e.g., CoLoR, InfLoRA)~\shortcite{color,inflora}, our approach differs by introducing a gradient-based regularization term specifically applied to the decomposition matrices within the NMT Transformer context, rather than applying general regularization to the full parameter set or using replay buffers.

This dual-faceted solution offers an efficient task-switching strategy that allows us to change the language and domain expertise of a given NMT model on the fly, enables interactive model adaptation to new contexts, and, through the regularization of task-independent low-rank matrices, we can incorporate new knowledge efficiently, while we mitigate the effects of the catastrophic forgetting phenomenon in NMT.

The key contributions of our approach are as follows:

\begin{itemize}
    \item \textbf{Efficient task-switching in NMT:} By introducing LoRA into the NMT architecture of our models, we were able to efficiently switch the language and domain expertise of our models. This was done using a fraction of the parameters, leading to significant savings in computational resources, and a performance on par with full-parameter strategies.
    \item \textbf{Interactive domain adaptation:} By employing a simplified Mixture of LoRA Experts with user-controllable scaling factors, we were able to adapt our NMT models to different domains and styles interactively, in a plug-and-play manner, and without the need for fine-tuning.
    \item \textbf{Low-rank and gradient-based regularization to efficiently tackle the Catastrophic Forgetting problem:} By implementing a low-rank and gradient-based regularization strategy specifically designed to address the catastrophic forgetting problem, we were able to incorporate new knowledge into our low-rank matrices while preserving most of the knowledge from the previous tasks (i.e. domains and new language-pairs).
\end{itemize}

One of the limitations of our approach is the trade-off between the rank of adaptation and efficiency. While a higher rank leads to better performance, it also reduces the parameter-efficiency with regard to full-parameter fine-tuning approaches. In addition to this, the calibration of multiple low-rank matrices requires careful consideration to prevent any single-domain dominance and ensure a balanced performance across multiple domains. Finally, while our regularization strategy may lead to slightly reduced performance on new tasks compared to a non-regularized approach, it proves its consistency in mitigating the catastrophic forgetting efficiently on previous tasks. Despite this, the proposed low-rank approach provides a solid foundation towards more adaptable, efficient, and robust continual learning NMT systems.

\section{Related work}\label{sec-related-work}

Continual learning has been an area of active research due to its potential to enable machine models to incrementally acquire, fine-tune, and transfer knowledge across tasks and domains without the need for periodic retraining~\shortcite{survey-cl}. However, the specific applications for NMT tasks have not been so widely studied~\shortcite{survey-nlp}. 

Previous works in continual learning for NMT~\shortcite{thompson-cf,few-shot-reg,luong-manning-2015-stanford} have primarily focused on strategies like rehearsal, regularization, and expansion, where models are either continuously trained on a mix of new and old data, regularized with regard to previous parameters, or expanded with new parameters for each task. Still, these methods often suffer from issues like the catastrophic forgetting phenomenon~\shortcite{mccloskey-cf,cf-cs-first,ratcliff} and inefficiency in parameter utilization~\shortcite{parameter-efficient-nlp,overparameterized-nn,lottery-ticket}. Similarly, other authors have proposed methods based on reinforcement learning~\shortcite{rl-nmt} or interactive online learning~\shortcite{peris2019online} approaches to tackle this problem. However, these solutions usually lack the efficiency needed to incorporate new knowledge dynamically without forgetting past information.

Our research introduces a novel application of the Low-Rank Adaptation method (LoRA) \shortcite{lora}, to mitigate the aforementioned issues within the context of NMT. This approach has been proven to be effective in maintaining high performance, while modifying only a minimal subset of model parameters, thus offering a solution to the parameter inefficiency problem.

The effectiveness of parameter-efficient fine-tuning (PEFT) methods has been increasingly recognized in the field. Approaches such as Adapter modules~\shortcite{parameter-efficient-nlp}, AdaLoRA~\shortcite{adalora}, Prefix Tuning~\shortcite{prefix-tuning}, P-Tuning~\shortcite{p-tuning}, or fine-tuning with sparse updates~\shortcite{sparse-fine-tuning} have shown that it is possible to achieve a high performance while only adjusting a small subset of the model's parameters. More recently, optimizations such as DoRA~\shortcite{dora-2024} and LoRA+~\shortcite{loraplus-2024} have further improved the convergence rates and expressive power of low-rank adapters, validating the continued relevance of this research direction. Recently, methods like CoLoR and InfLora have explored LoRA in continual learning settings~\shortcite{color,inflora}. While CoLoR focuses on continual few-shot learning for generative tasks and InfLora targets the specific mechanics of SVD updates, our work focuses specifically on the application to dedicated NMT architectures. Unlike standard EWC which regularizes the full parameter set~\shortcite{ewc}, or adapter-based methods that introduce additional layers, our method introduces a regularization term specifically for the $X$ and $Y$ decomposition matrices. This allows for lighter-weight storage and task-switching compared to storing full Fisher Information Matrices for the entire network.

The continual domain adaptation problem in NMT has been investigated to a lesser extent, and the usage of multiple low-rank matrices for domain adaptation is a relatively new and unexplored terrain~\shortcite{survey-cl,survey-nlp,lora}. For instance, some authors showed that linear mixtures are better for combining domain-specific probabilities~\shortcite{mixture-models}. Other methods for combining aspects of different domains have been explored from multiple perspectives~\shortcite{multi-domain-nmt,ensembles-nmt}, but without specifically leveraging the efficiency of low-rank approaches. While multi-LoRA composition and scalable adapter serving have been explored in computer vision~\shortcite{multi-lora} and Large Language Models~\shortcite{slora-2024}, our approach extends these concepts to NMT by applying a linear combination of LoRAs without gating networks, enabling real-time user-driven style control.

The regularization of low-rank matrices in the context of continual learning represents a promising direction to efficiently tackle the catastrophic forgetting problem. While regularization techniques such as L2, Learning without Forgetting (LwF)~\shortcite{lwf}, and Elastic Weight Consolidation (EWC)~\shortcite{ewc} have been well-studied in continual learning, among many others~\shortcite{cf-da-progressive-networks,cf-da-neurogenesis,cf-gr-fearnet}, their application to low-rank matrices in NMT is new~\shortcite{llm-mt-lora,cl-nmt,cf-survey}. These regularization techniques usually work by penalizing changes on parameters that are important for past tasks, so applying these strategies directly into low-rank matrices presents a unique opportunity to prevent the catastrophic forgetting phenomenon more efficiently, with handy task-switching capabilities at no cost. In contrast, our work is among the first to explore this intersection, focusing on the ability of low-rank and gradient-based regularization to preserve linguistic knowledge over time in NMT models~\shortcite{llm-mt-lora,cf-survey}.

Therefore, while there are various strategies for continual learning in NMT~\shortcite{llm-mt-lora,cl-nmt}, this work differentiates itself by focusing on the novel application of low-rank matrices, which allows efficient task-switching strategies, dynamic and interactive domain adaptations, and the ability to incorporate new knowledge into the models by addressing the catastrophic forgetting problem through a specifically designed regularization strategy. This work presents itself as a more nuanced and computationally efficient approach to the continual learning problem in NMT than what has previously been explored in the literature.

\section{Methodology}\label{sec-methodology}

The continual learning problem in Neural Machine Translation (NMT) can be framed as an optimization problem in which the goal is to minimize a loss function $L$ over a sequence of tasks $T_1, T_2, \ldots, T_N$ (such as different languages, domains, and styles), while preserving most of the performance from previously learned tasks. Formally, this concept can be represented as:

\begin{equation}
    \operatorname*{arg\,min}_{\theta} \sum_{n=1}^{N} L(T_n, f_{\theta_n})
\end{equation}
where $f_{\theta_n}$ denotes the NMT model parameterized by $\theta_n$ after learning task $T_n$.

The primary challenge lies in updating $\theta_n$ for each new task $T_n$ in such a manner that any increase in loss for earlier learned tasks is minimal.

To do so, we address this issue by leveraging strategies centered around: 1) Parameter regularization, which is particularly effective in heavily over-parameterized neural networks where redundant parameters are identifiable; 2) Dynamic architectures, which expand the parameters of a base model to accommodate new tasks; 3) Rehearsal strategies, which combine previous information along with new data to mitigate the catastrophic forgetting effects, until the network is saturated. 

As each strategy tries to mitigate the catastrophic forgetting problem from a distinct angle, but without accounting for other issues such as data availability, model saturation, and model efficiency, we propose a unified solution: Gradient-based regularization for low-rank matrices. This dual-faceted approach includes:
\begin{itemize}
    \item Parameter-Efficient Fine-Tuning (PEFT) for addressing parameter-efficient fine-tuning issues and network saturation problems.
    \item Regularization method to mitigate the catastrophic forgetting problem.
\end{itemize}

The PEFT method aims to reduce the computational cost and parameter count needed for task-specific fine-tuning while decoupling task-specific parameters from the base model. In this work, we focus our attention on the Low-Rank Adaptation (LoRA) method~\shortcite{lora}, which instead of inefficiently storing all the weights of the network after fine-tuning, freezes part of it and learns the low-rank matrices needed to approximate the expected high-dimensional weight matrices of the fine-tuned neural network in a much more efficient way. In other words, in a traditional fine-tuning approach we would typically start with a weight matrix $W$ and after the fine-tuning process, we would end up with a weight matrix $W'$. Therefore, this means that the changes in this matrix $W$ could be represented by a second matrix ${\Delta W}$ so that the final matrix $W'$ could be computed like this:
\begin{equation}
    W' = W + {\Delta W}
\end{equation}

With this in mind, LoRA proposes a matrix factorization approach to approximate this weight matrix \( \Delta W \in \mathbb{R}^{p \times q} \) by two lower-rank matrices \( X \) and \( Y \), such that \( \Delta W \approx XY^\top \). As a result, the LoRA-based update is formulated as:
\begin{equation}
    W' = W + XY^\top; \quad W'_{i,j} = W_{i,j} + \sum_{k=1}^{r} x_{ik} y_{jk}
\end{equation}
where $X \in \mathbb{R}^{p \times r}$ and $Y \in \mathbb{R}^{q \times r}$ are the low-rank matrices, with $r \ll \min(p, q)$ being the rank of the adaptation, and $i, j, k$ represents the element in the i-th or jth row and k-th column of the matrix.

This decomposition reduces the parameter space from \( pq \) to \( r(p + q) \), significantly reducing the computational complexity. Hence, this facilitates efficient fine-tuning and task-switching capabilities in NMT models, enabling efficient model adaptation to different languages and domains by integrating these task-specific low-rank components \( XY^\top \) into the base model. Therefore, for task-specific adaptations, the model weights are updated as follows:

\begin{equation}
    W'_{T_n} = W + (XY^\top)_{T_n}
\end{equation}

Once these matrices are trained independently for various tasks (e.g. domains or styles), we can perform interactive model adaptations through a calibrated linear combination of multiple LoRA matrices\footnote{During our testing we had to calibrate (reweight) the individual contributions of each LoRA with a domain-specific coefficient as not all domains had the same impact in the final model's performance.}, akin to a simplified Mixture of LoRA Experts (MoLE) but without a gating network, aiming to enhance the usability and to be able to interactively bias the model's knowledge along desired dimensions without the need for fine-tuning. Unlike traditional MoEs that rely on networks to weigh contributions, we allow domain-specific coefficients and a user scaling factor ($\alpha$) for fine-grained control. Consequently, the adaptation is defined as follows:

\begin{equation}
    W'_{MoLE} = W + \sum_{n=1}^{N} \alpha_n \lambda_n X_n Y_n^\top
\end{equation}
where \( N \) is the number of tasks (i.e. domains or styles), \( \lambda_n \) are domain-specific coefficients, \(\alpha_n\) is a user-adjustable scaling factor, and \( X_n, Y_n\) are the respective LoRA matrices, all of them for their corresponding task $T_n$. This calibration ensures a balanced performance across domains, preventing any single domain from compromising the performance of the rest. In our experiments, the coefficients $\lambda$ are set heuristically to achieve the desired blend of domains.

\begin{figure}[ht]
\centering
\vspace*{-0.65cm}
\includegraphics[width=0.75\textwidth]{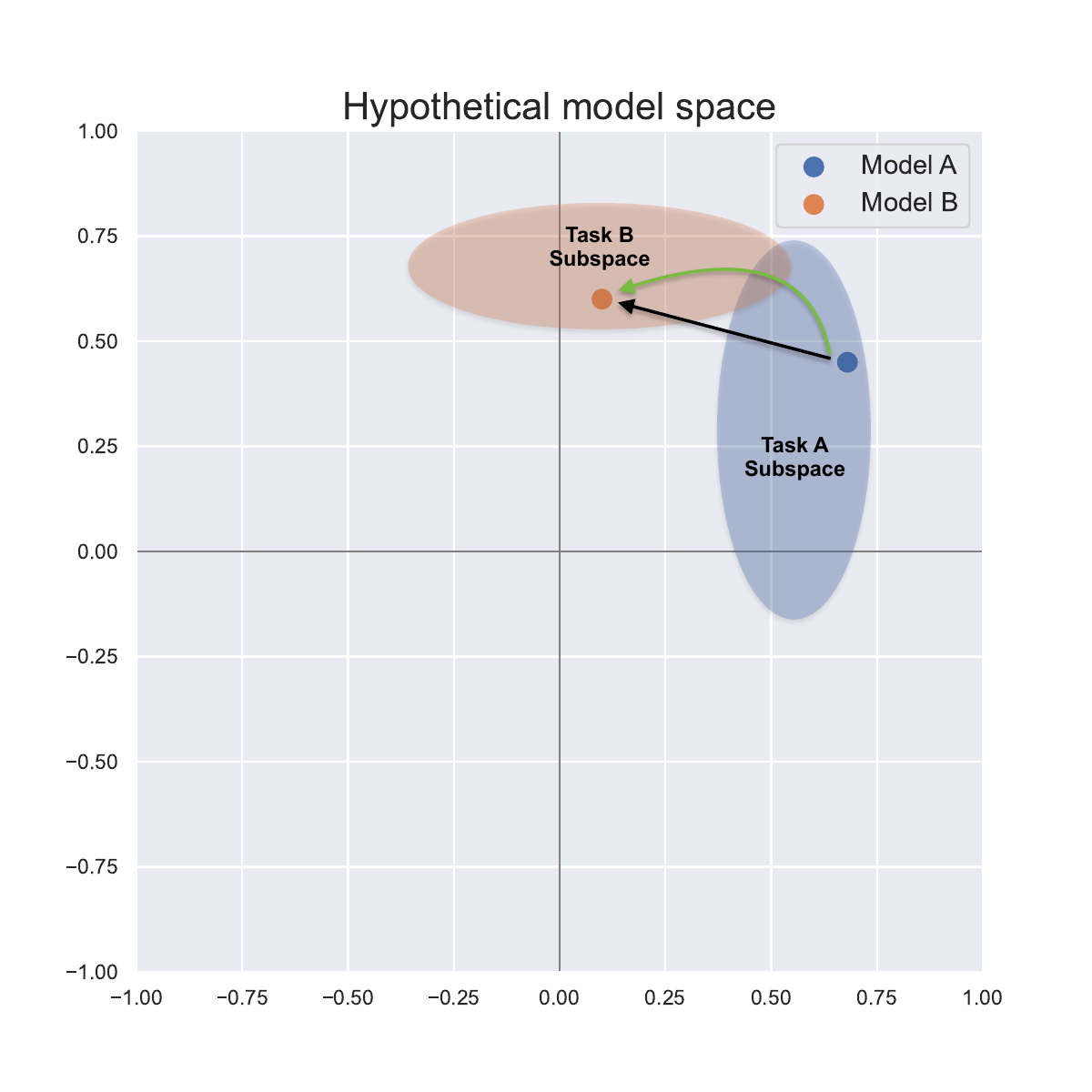}
\caption{\textbf{Theorized model space}: If we consider a model as a point in a hyperdimensional space resulting from the data distribution on which it was trained, we can expect that another model trained on a different data distribution, but similar, could end up in the shared region defined by these subspaces. Therefore, to mitigate the catastrophic forgetting phenomenon we need to find a way to move the model into that overlapping region with the available information. (The arrows indicate the potential path for the $f_{A, B}$ model; in green the arrow that passes across the overlapping area, and in black the arrow that skips this area).}\label{fig:model-space}
\end{figure}

Finally, to address the problem of catastrophic forgetting and to be able to incorporate new knowledge into our low-rank matrices, we considered that a pre-trained model $f_{\theta}$ could be viewed as a point in a hyperdimensional space, defined by its matrices and vectors, which are the result of the training data distribution in which it has been trained on. Consequently, we hypothesized that since a task defines a region in that space, two similar tasks should have an overlapping area at some point (See Figure~\ref{fig:model-space}). Therefore, the problem of mitigating the catastrophic forgetting phenomenon becomes finding a way to move our model into the subspace region shared by both tasks. 

Assuming such an overlap exists between tasks $A$ and $B$, we can attempt to adjust the model weights to move them into this shared region. Conversely, if tasks are non-overlapping, we rely on the $\Delta W$ transformation inherent to LoRA to map one model state to another.

However, given the fact that in our problem definition, we only have access to the weights of the previous models and the data for the new task, we cannot simply merge these datasets to find this region, so we decided to take linear algebra approach by penalizing deviations in direction and magnitude with regard to the previous model.

Formalizing the previous concept, given a model $f_{\theta}$ trained on domain $D_A$, resulting in a model $f_{A}$\footnote{$f_{A}$, $f_{B}$, $f_{A, B}$ are used instead of $f_{\theta_A}$, $f_{\theta_B}$, $f_{\theta_{A, B}}$ to simplify the notation.}, we aim to adapt this model to a new domain $D_B$ while minimizing the performance degradation on the original domain $D_A$, obtaining a new model $f_{A, B}$. 

Consequently, if we represent these two models, conceptualized as points in a hyperdimensional space by a set of vectors, $W_{A}$ and $W_{A, B}$, we could transform one into the other by simply calculating the vector that represents the difference between $W_{A}$ and $W_{A, B}$, and applying it to $W_{A}$. However, since we don't know where this $W_{A, B}$ point will be located, instead, we added a penalty term to our objective function to control the magnitude (distance) with which to prevent the new weights ($W_{A, B}$) from drifting too far from the previous ones ($W_{A}$). Specifically, we decided to use the norm of their difference as our distance function, and then, we raised it to a tunable power, $\left \| W_{A,B} - W_{A} \right \|^\gamma$, to have greater control over the regularization effects. Similarly, the direction in which $W_{A,B}$ should be moving to become closer to $W_{B}$, is defined by the domain $D_B$ (dataset). However, if we go blindly in this direction, from one point to another, we can easily escape the overlapping region that would allow us to have a $W_{A,B}$ model (See Figure~\ref{fig:model-space}). To prevent this, we adjusted (re-weighted) the new weight updates by factoring in the importance of weights from the previous task, similar to pruning strategies~\shortcite{acc-grad-1,acc-grad-2}. We calculated this importance using the cumulative gradient of the previous model, $f_{A}$, after its training, computed as follows:

\begin{equation}
G_{\theta} = \frac{1}{M}\sum_{i=1}^{M} \nabla L(f_{\theta}, x_i, y_i) 
\end{equation}
where $G_{\theta}$ is the accumulated gradient for the training data (or a representative subset), $\nabla L(f_{\theta}, x_i, y_i)$ represents the gradient of the loss function $L$ with respect to the NMT model, $f_{\theta}$, parameterized by $\theta$ (where $\theta$ refers to the full model parameters), for a single training example $(x_i, y_i)$. $M$ represents the total number of datapoints (sentences) in the previous task's dataset used to compute this importance metric.

From all these components, we introduced a gradient-based regularization specifically designed for our low-rank matrices, altering the typical loss function as follows:
\begin{equation}
\begin{split}
    \mathcal{L}'_K &= \mathcal{L}(X_K, Y_K) + \lambda_{reg} \sum_{n=1}^{K-1} 
\Bigl[
    G_{X, n} \bigl\lVert (X - X_n) \bigr\rVert^\gamma 
    \;+\; 
    G_{Y, n} \bigl\lVert (Y - Y_n) \bigr\rVert^\gamma
\Bigr]
\end{split}
\end{equation}
where $\mathcal{L}$ is the primary task loss, $X_n$ and $Y_n$ are the low-rank matrices learned for task $T_n$, $\lambda_{reg}$ is a regularization coefficient, $\gamma$ is an exponential coefficient, and $G_{X, n}$ and $G_{Y, n}$ represent the cumulative gradients for task $T_n$ specifically with respect to parameters $X$ and $Y$.\footnote{It is worth to point out that, if we purposefully ignore the gradient, we may end up with an L1 or L2 regularization for the low-rank matrices, depending if we set $\lambda$ equal to $1$ or $2$.} The coefficients $\lambda_{reg}$ and $\gamma$ were set via a grid search on a hold-out validation set to optimize the harmonic mean between stability (old task) and plasticity (new task), as detailed in the experimental section. Consequently, the optimization problem for the low-rank matrices is defined as follows:

\begin{equation}
(X_n, Y_n) = \operatorname*{arg\,min}_{X, Y} L'_n(X,Y)
\end{equation}
where the objective is to find the optimal low-rank matrices, $X_n$ and $Y_n$, that minimize the modified loss function $L'_n(X,Y)$.\footnote{Given the fact that there are multiple low-rank matrices per model, our $(X, Y)$ notation is a simplification that comprehends all of them.}

Hence, through this holistic approach, we have introduced an efficient and nuanced solution for adapting NMT models to new tasks (languages, domains, and styles) either by employing a parameter-efficient task-switching strategy, interactively adapting their domain expertise, and finally, by incorporating new knowledge into these low-rank matrices using a specifically designed method to mitigate the effects of the catastrophic forgetting phenomenon in this scenario.

\section{Experimentation and Results}
\label{sec-results}

\subsection{Experimental Setup}\label{sec-exp-setup-new}

This section elaborates on the experimental setup for our study on tackling the continual learning problem in neural machine translation (NMT) using low-rank adaptation (LoRA) techniques. We outline the datasets used, evaluation metrics, and implementation details, providing a comprehensive view of the experimental environment.

\subsubsection{Dataset Description}

The datasets selected for this work are diverse in nature, encompassing multiple languages, domains, and stylistic variances (See Table~\ref{tab:datasets}).

\begin{table}[h!]
\centering
\caption{\textbf{Datasets:} To control for potential domain overexposure bias and accelerate our experimentation, we employed reduced dataset versions (approximately 100,000 sentences per dataset, 30,000 for Multi30k). Details are provided in Subsection~\ref{sec-detailed-results}.}\label{tab:datasets}
\resizebox{\textwidth}{!}{%
\begin{tabular}{@{}llll@{}}
\toprule
\textbf{Dataset Type}            & \textbf{Dataset Name} & \textbf{Domains}           & \textbf{Language Pairs} \\ \midrule
\multirow{1}{*}{Multi-Language}     & Europarl                       & EU proceedings            & \begin{tabular}[c]{@{}l@{}}English-[Spanish, French, \\ German, Czech]\end{tabular}  \\ \midrule
\multirow{3}{*}{Domain-Specific} & SciELO                & Scientific-Health         & English-Spanish         \\ \cmidrule(l){2-4} 
                                 & SciELO                & Scientific-Biological     & English-Spanish         \\ \cmidrule(l){2-4} 
                                 & JRC-Acquis            & Legal                     & English-Spanish         \\ \midrule
\multirow{1}{*}{Stylistic Variance} & Multi30k-Formality             & Neutral, Formal, Informal & English-Spanish                                                \\ \bottomrule
\end{tabular}
}
\end{table}

Getting into the details of these datasets, we describe them as follows:

\begin{itemize}
    \item \textbf{Multi-Language Datasets:} Used to study the limits of our approach, taking into account that the less similar two given languages are, the more aggressive the effects of catastrophic forgetting are expected to be.
    \item \textbf{Domain-Specific Datasets:} The purpose of these datasets was to study the inter-domain dependencies for a given language in relation to the continual domain problem.
    \item \textbf{Stylistic Variance Datasets:} A synthetically generated set of datasets to study the effects of style in an isolated manner, to avoid language-specific or domain-related noise. While synthetic data may not perfectly reflect real-world linguistic noise, this setup was necessary to isolate the effect of the decoder's adaptation. By keeping the source sentences identical while varying only the target formality, we ensure that performance differences are attributable to the LoRA adaptation rather than encoding discrepancies.
\end{itemize}

\subsubsection{Evaluation Metrics}

The performance of our approach was evaluated primarily using the BLEU metric from the sacreBLEU toolkit~\shortcite{sacrebleu}, which is the reference BLEU implementation for the WMT conference~\shortcite{wmt16}. In addition to this, we also used the character n-gram F-score (chrF++)~\shortcite{metric-chrf} and the Translation Error Rate (TER)~\shortcite{metric-ter}, also included in the sacreBLEU toolkit. However, no major differences were found between these metrics during the evaluation. We acknowledge that relying solely on automated metrics such as BLEU provides an incomplete picture of the model's fluency and adequacy compared to human evaluation. However, for the purpose of this comparative study across a large number of runs and domains, automated metrics provide a consistent standard for relative improvement.

\subsubsection{Implementation Details}

For the implementation of our NMT models, the following setup was used:
\begin{itemize}
\item \textbf{Model Architecture and Training Environment:}
    \begin{itemize}
        \item \textbf{Base Architecture:} Transformer architecture~\shortcite{transformer} was used as the foundation for our NMT models. These models were constructed with approximately 10 million parameters. Each model featured 3 layers, 8 attention heads, an embedding dimension of 256, and a feedforward layer dimension of 512.
        \item \textbf{Computational Resources:} Training was conducted on a system equipped with an RTX 4090 GPU (24GB VRAM), supported by a 13th Gen Intel(R) Core(TM) i7-13700K CPU, and complemented with 64GB of RAM. The operating system was an Ubuntu 22.04 LTS with CUDA Version 12.0.
    \end{itemize}
\item \textbf{Training Process and Parameters:}
    \begin{itemize}
        \item \textbf{Training Tool:} AutoNMT~\shortcite{autonmt}, a tool that streamlines the research of seq-to-seq models, by automating the preprocessing, training, and evaluation of NMT models. Internally, AutoNMT utilizes PyTorch 2.0.
        \item \textbf{Hyperparameters:} We employed cross-entropy loss (without label smoothing) and AdamW as the default optimizer. Our training batches consisted of 128-1024 sentences, depending on the model requirements. The maximum token length was established between 150-400, varying depending on the dataset. The clip-norm was fixed at 1.0. The domain specific coefficients $\lambda_n$ and user scaling factor $\alpha_n$ for the mixture experiments were determined heuristically, while the regularization coefficients were determined via grid search.
        \item \textbf{Training Duration:} Models were trained for a maximum of 50-200 epochs, with an early stopping implemented (patience set at 10-15 epochs) to prevent overfitting.
    \end{itemize}
\end{itemize}

\subsection{Results}
\label{sec-detailed-results}

In this section, we present our experimental findings. We structure the results to first establish the baseline effectiveness of LoRA for NMT (Section~\ref{subsec-lora-ft}), then explore its composability for interactive adaptation, and finally demonstrate the proposed method to mitigate catastrophic forgetting. To provide a comprehensive analysis, we employed several metrics, including chrF++~\shortcite{metric-chrf} and TER~\shortcite{metric-ter}. However, for brevity reasons and because there were no differences in our findings, this section focuses mainly on reporting the BLEU~\shortcite{sacrebleu} score results.

\subsubsection{Effectiveness of Low-Rank Adaptation (LoRA) for Parameter-Efficient NMT fine-tuning}
\label{subsec-lora-ft}

This subsection delves into the effectiveness of Low-Rank Adaptation (LoRA) for fine-tuning NMT models in a parameter-efficient manner. To do so, we decided to explore two different scenarios: 1) Domain-Specific model adaptation and, 2) Efficient language-switching.

\begin{figure}[ht]
\centering
\includegraphics[width=1.0\textwidth]{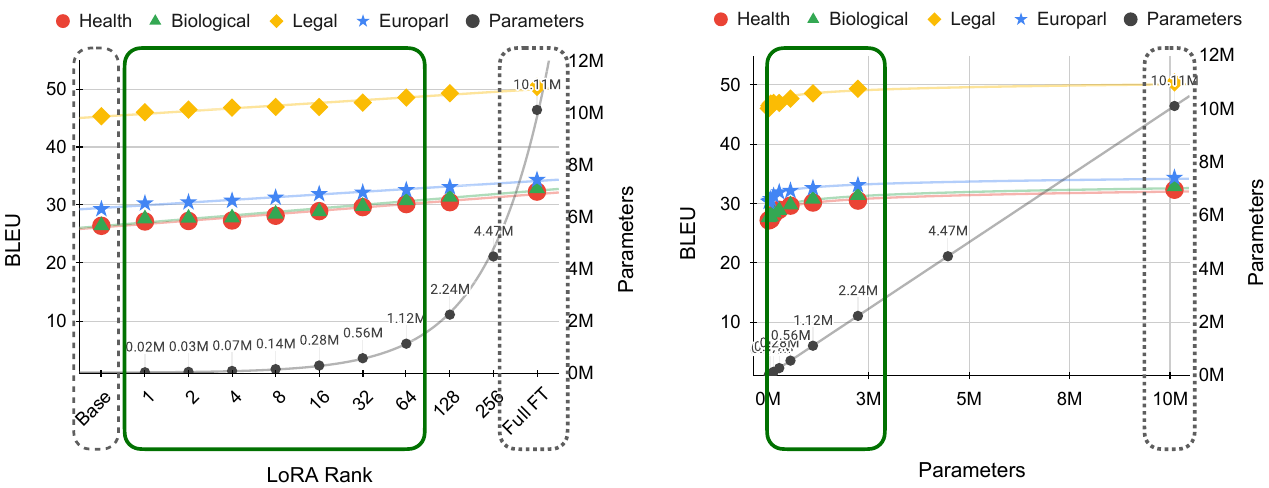}
\caption{\textbf{Comparative analysis of the model performance w.r.t the LoRA-rank and the number of parameters used:} This figure shows the performance improvements (BLEU scores) achieved through a low-rank fine-tuning strategy across various domains (i.e. Health, Biological, Legal, and Europarl). As we can see, the LoRA fine-tuning strategy defines a regime of logarithmic performance improvements so that we can use a minimal fraction of parameters to achieve a performance similar to a traditional full-parameter fine-tuning. Note: M = millions.}
\label{fig:lora-domains}
\end{figure}

For our first experiment, we decided to train a generic machine translation model on a variety of domains by sampling a balanced mixture of sentences from our English-Spanish datasets. This controlled mixed-domain dataset reduced potential biases from overexposure to any single domain and provided a distinct baseline. Next, we fine-tuned this model using a low-rank fine-tuning strategy for a wide range of ranks (1 to 256), going from 17,472 to 2,236,416 parameters, in order to study the performance improvements in relation to the number of parameters used during the training (See Figure~\ref{fig:lora-domains}). In addition to this, we trained one additional model per domain using the traditional full-parameter fine-tuning approach to serve as a comparison of the number of parameters that this approach needs with regard to a low-rank approach. Finally, we added the baseline results of our pre-trained model (no fine-tuning) for each domain as the starting point.

As we can see from Figure~\ref{fig:lora-domains}, which shows the linear (left) and logarithmic (right) performance improvements with regard to the number of parameters used during fine-tuning, we observed that as we increase the number of parameters during fine-tuning, we get a logarithmic increase in the performance of our model, regardless of the domain. That is, the smaller LoRA ranks gave us exponential improvements in performance (right figure) and marginal parameter growth, while the larger ranks gave us linear improvements of improvements but a critical exponential growth of these parameters. 


Similarly, in Table~\ref{tab:lora-params} we can see the individual performance improvements per domain as well as the average improvement as a function of the LoRA-rank and the parameters used. Again, we use the full-parameter fine-tuning model as a reference. From these results, we can observe that even with 1-5\% of the parameters, we can achieve on average between 30 to 50\% of the fine-tuning performance, and even with a negligible 0.17\% of the parameters, we were able to achieve up to a 16.20\% of the fine-tuning performance (See Table~\ref{tab:lora-params}). Likewise, with a rank of 64, we achieve a 65.19\% of the full-parameters performance, but using just 11.06\% of its parameters.\footnote{Rank 1: 3.04/18.76; Rank 64: 12.23/18.76}

\begin{table}[ht]
\centering
\caption{\textbf{Relative performance improvements per domain:} Smaller LoRA ranks allowed us to achieve performance improvements close to their full-parameter counterparts at a fraction of their parameters. Note: K = thousands, M = millions.}
\label{tab:lora-params}
\resizebox{\textwidth}{!}{%
\begin{tabular}{@{}lrrrrrr@{}}
\toprule
\textbf{}    & \multicolumn{1}{l|}{\textbf{}} & \multicolumn{5}{c}{\textbf{Domain boost w.r.t. base model}}                                             \\ \cmidrule(r{0.25em}){1-2} \cmidrule(l{0.25em}){3-7} 
\textbf{LoRA-Rank} &
  \multicolumn{1}{c|}{\textbf{Params used}} &
  \multicolumn{1}{c}{\textbf{Health}} &
  \multicolumn{1}{c}{\textbf{Biological}} &
  \multicolumn{1}{c}{\textbf{Legal}} &
  \multicolumn{1}{c|}{\textbf{Europarl}} &
  \multicolumn{1}{c}{\textbf{Average}} \\ \midrule
\textbf{1}   & \multicolumn{1}{r|}{17.4K (0.17\%)}       & 2.82\%  & 4.44\%  & 1.50\%  & \multicolumn{1}{r|}{3.39\%}  & 3.04\% $\pm$ 1.23\%  \\
\textbf{2}   & \multicolumn{1}{r|}{34.9K (0.35\%)}       & 3.17\%  & 4.87\%  & 2.49\%  & \multicolumn{1}{r|}{4.04\%}  & 3.64\% $\pm$ 1.04\%   \\
\textbf{4}   & \multicolumn{1}{r|}{69.8K (0.69\%)}       & 3.48\%  & 5.08\%  & 3.27\%  & \multicolumn{1}{r|}{4.99\%}  & 4.20\% $\pm$ 0.96\%  \\
\textbf{8}   & \multicolumn{1}{r|}{139.7K (1.38\%)}      & 6.59\%  & 7.53\%  & 3.54\%  & \multicolumn{1}{r|}{6.85\%}  & 6.13\% $\pm$ 1.77\%  \\
\textbf{16}  & \multicolumn{1}{r|}{279.5K (2.77\%)}      & 9.60\%  & 9.09\%  & 3.50\%  & \multicolumn{1}{r|}{8.93\%}  & 7.78\% $\pm$ 2.87\%  \\
\textbf{32}  & \multicolumn{1}{r|}{559.1K (5.53\%)}      & 12.25\% & 12.55\% & 5.13\%  & \multicolumn{1}{r|}{9.80\%}  & 9.93\% $\pm$ 3.43\%  \\
\textbf{64}  & \multicolumn{1}{r|}{1.1M (11.06\%)}   & 14.22\% & 16.33\% & 7.09\%  & \multicolumn{1}{r|}{11.27\%} & 12.23\% $\pm$ 4.00\% \\
\textbf{128} & \multicolumn{1}{r|}{2.2M (22.13\%)}   & 15.47\% & 18.69\% & 8.78\%  & \multicolumn{1}{r|}{13.04\%} & 14.00\% $\pm$ 4.18\% \\ \midrule
\textbf{FT}  & \multicolumn{1}{r|}{10.1M (100.00\%)} & 22.40\% & 24.24\% & 11.15\% & \multicolumn{1}{r|}{17.25\%} & 18.76\% $\pm$ 5.88\% \\ \bottomrule
\end{tabular}
}
\end{table}

To further explore the effectiveness of low-rank adaptation strategies for parameter-efficient fine-tuning, we decided to repeat the previous experiment but this time fine-tuning only the decoder. However, due to the lack of computational resources with which to train a large NMT model capable of generating high-quality encodings that were truly independent of the target sentences with which the source sentences were paired, we decided to generate a synthetic dataset in which the source sentences were always the same (thus, minimizing the encoding noise), while the target sentences were available at different levels of formality. 

To do this, we first trained an NMT model from scratch on the Multi30K dataset, which we consider to have a neutral level of formality. Then, we fine-tuned the decoder of this pre-trained model, for a wide range of LoRA ranks, on the Multi30k-Formal and Multi30k-Informal datasets. Again, we fine-tuned the full-parameter model as a reference. As a result, all the datasets share the same source sentences in this experiment, and all the models share the same encoder, we were able to isolate and study the effects of the low-rank fine-tuning method just on the decoder. 

These results can be seen in Table~\ref{tab:multi30k}, where the target domain always outperforms all other domains even when only 8,736 parameters (rank 1) were used during training. This is especially important because it means that by simply storing 17KB~\footnote{8,736 parameters x 16 bits} in 16-bit precision, we can add new capabilities on the fly for an NMT model that weights 40MB, thus saving a considerable amount of storage.

\begin{table}[ht]
\centering
\caption{BLEU results from the Multi30k-Formality dataset (Neutral, Informal, and Formal domains).}
\label{tab:multi30k}
\begin{tabular}{@{}lrrrrrrr@{}}
\toprule
& \multicolumn{1}{l}{} & \multicolumn{3}{@{}c@{}}{\textbf{Neutral→Informal}} & \multicolumn{3}{@{}c@{}}{\textbf{Neutral→Formal}}  \\ \cmidrule(lr{0.25em}){3-5} \cmidrule(l{0.25em}){6-8}%
\textbf{Rank}           & \textbf{Parameters}   & \textbf{Neutral}  & \textbf{Informal}  & \textbf{Formal} & \textbf{Neutral} & \textbf{Informal} & \textbf{Formal} \\ \midrule
\textbf{1}  & 8,736   & 49.98 & \textbf{50.66} & 50.05 & 50.29 & 50.32 & \textbf{50.31} \\
\textbf{2}  & 17,472  & 50.15 & \textbf{50.77} & 50.44 & 50.38 & 50.38 & \textbf{50.43} \\
\textbf{4}  & 34,944  & 49.97 & \textbf{50.70} & 50.17 & 50.39 & 50.53 & \textbf{50.76} \\
\textbf{8}  & 69,888  & 50.16 & \textbf{50.97} & 50.44 & 50.67 & 50.84 & \textbf{51.09} \\
\textbf{16} & 139,776 & 50.14 & \textbf{51.01} & 50.37 & 50.80 & 50.95 & \textbf{51.31} \\
\textbf{32} & 279,552 & 50.42 & \textbf{51.27} & 50.71 & 50.84 & 51.14 & \textbf{51.33} \\
\textbf{64} & 559,104 & 50.66 & \textbf{51.36} & 50.80 & 51.35 & 51.59 & \textbf{51.73} \\ \midrule
\textbf{FT} & 10,176,576 & 51.24 & \textbf{51.60} & 51.40 & 50.72 & 51.47 & \textbf{51.94} \\ \bottomrule
\end{tabular}
\end{table}

\begin{figure}[ht]
\centering
\includegraphics[width=1.0\textwidth]{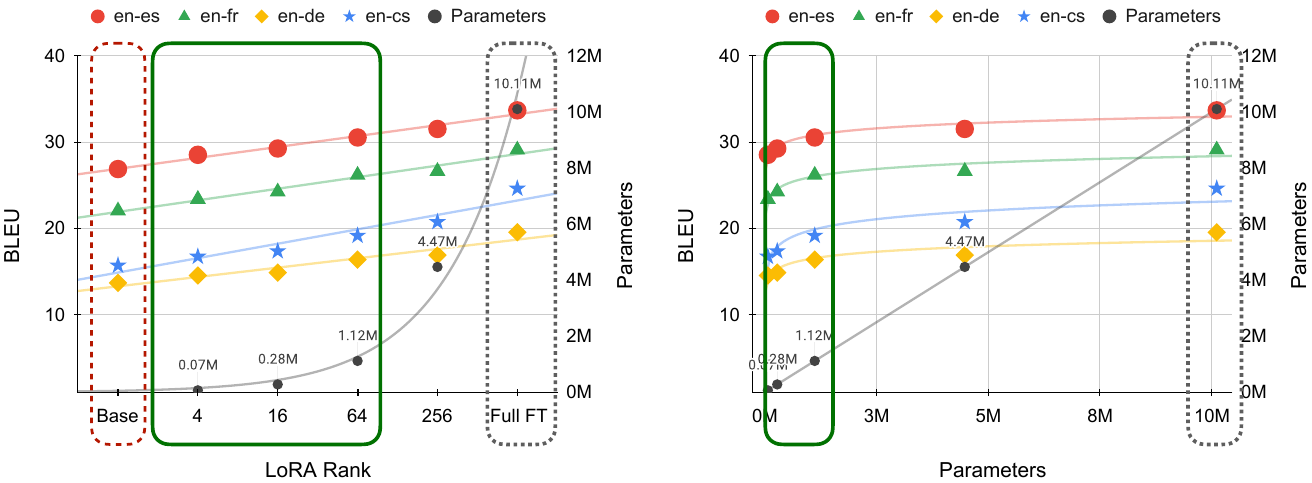}
\caption{\textbf{Efficient performance boosting for multilingual NMT models:} This figure demonstrates the effectiveness of low-rank adaptations in improving the performance of a multilingual NMT model trained on limited data (25k sentences per language). The BLEU score improvements for each language pair show significant gains with minimal parameters, following a logarithmic trend until the performance matches the full-parameter fine-tuning approach. (M = millions).}
\label{fig:lora-langs-pretrained}
\end{figure}

\begin{figure}[ht]
\centering
\includegraphics[width=1.0\textwidth]{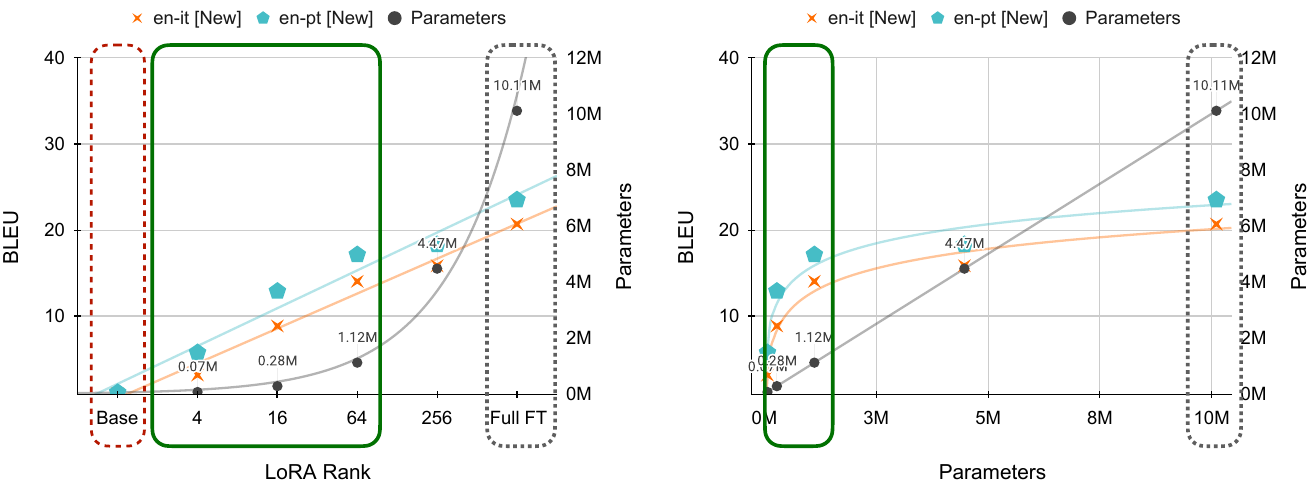}
\caption{\textbf{Incorporating new unseen languages into a pre-trained multilingual NMT model:} This figure illustrates the effectiveness of extending a multilingual NMT model to include new language-pairs (en-it and en-pt), completely from scratch, in a parameter-efficient manner, thus, demonstrating consistent performance improvements with an increased parameter utilization. (M = millions).}
\label{fig:lora-langs-new}
\end{figure}

After this experiment, we decided to test whether low-rank fine-tuning strategies could be used to switch the language of a given pre-trained model using a minimal amount of parameters. To do so, we designed two experiments: The first one, consisted of training an NMT model using a minimal amount of data from several languages, as a warmup, with the goal of helping the model to discriminate between different languages more easily and be able to later boost the performance in that language using our low-rank strategy (See Figure~\ref{fig:lora-langs-pretrained}). Then, for the second experiment, we tried to learn a different set of new languages completely from scratch, with the goal of studying if the previous technique was only valid for boosting a known language or indeed, it could be used to add new languages to a pre-trained model efficiently (See Figure~\ref{fig:lora-langs-new}).

In Figure~\ref{fig:lora-langs-pretrained} we can see the results from the first experiment, where we first pre-trained a base multilingual model using 25k sentences on four language-pairs (English-Spanish, English-French, English-German, English-Czech), and then, we boosted the knowledge expertise for each individual language using our proposed low-rank strategy. From these results, we can observe that as it happened in the previous domain experiment (See Figure~\ref{fig:lora-domains}), we can effectively increase the performance on a given language up to 80-90\% of the performance of the full-parameter approach but using just a 10\% of its parameters. Again, this function follows a logarithmic trend, so when enough parameters are used, it matches the performance of a traditional full-parameter fine-tuning approach.


Next, we decided to move this experiment one step further and test whether we could incorporate new languages, completely unseen during pre-training, into our previous multilingual NMT model in a parameter-efficient manner. To do so, we trained a wide range of low-rank matrices for two new language-pairs (English-Italian and English-Portuguese), and even though it could be argued that we could have used a more different set of languages, such as Russian or Chinese, we were limited to Romanic and Germanic languages due to our vocabulary choice (tokenizer) in this work, as it had been only trained on those languages. Nevertheless, given the wide range of languages and domains used during our experimentation, we expect these results to generalize to other languages as well. 

Accordingly, in Figure~\ref{fig:lora-langs-new} we can observe the consistency of this approach, where the new languages are successfully learned in a parameter-efficient manner, following a logarithmic trend as we increase the number of parameters used in our low-rank approach until the results match the ones from the full-parameter fine-tuning. In addition to this, the parameter-performance trends from learning these new languages are consistent with the trends from the language-boosting experiment (See Figure~\ref{fig:lora-langs-pretrained}), although it is clear that due to the complexity of learning a new language, higher ranks are needed when compared to learning a new domain from scratch. Nonetheless, it can be clearly seen in the right Figure~\ref{fig:lora-langs-new} that, with a rank of 64, we got a 72.94\% of the performance of the full-parameter fine-tuning approach for the English-Portuguese pair, but using just 11.06\% of its parameters. Similarly, these stats are consistent with the English-Italian pair.


\begin{figure}[ht]
\centering
\includegraphics[width=0.75\textwidth]{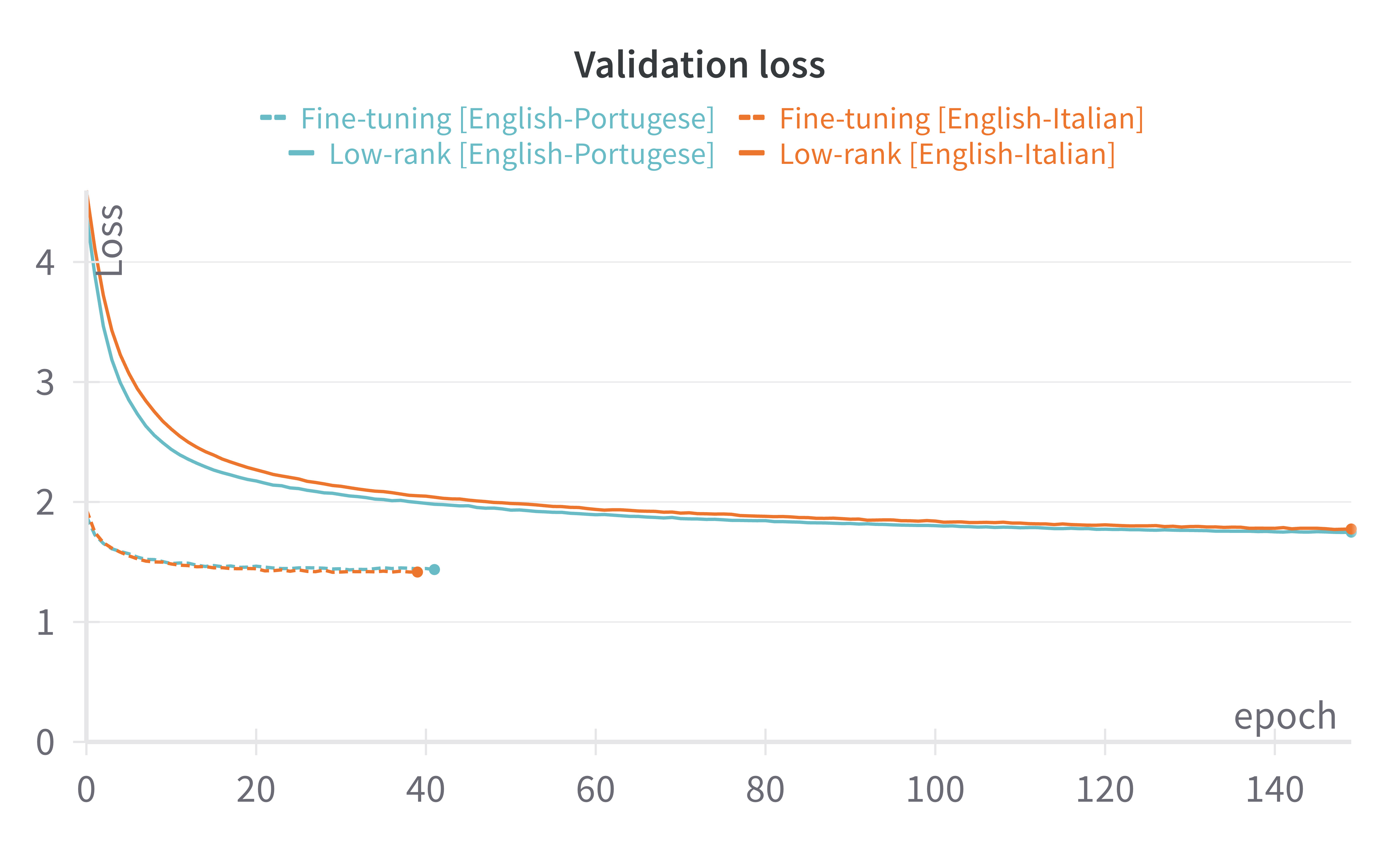}
\caption{\textbf{Comparative analysis of training times for low-rank and full-parameter fine-tuning approaches:} The low-rank approach is characterized by its longer training times (solid lines) when compared to a full-parameter approach (dashed lines) due to the complexity of capturing complex nonlinear dependencies through a matrix factorization process. In this case, the low-rank approach used 44.26\% (rank 256) of the full-parameter model so higher loss and training times are expected.}\label{fig:convergence-rate}
\end{figure}

One of the limitations of this approach is that, since we are learning the low-rank components that factorize the weight matrices of the desired fine-tuned model, the adaptation process may be less effective in capturing complex nonlinear dependencies compared to the full-parameter fine-tuning approach, potentially leading to suboptimal performance in tasks that require intricate feature interactions or when we have to deal with highly variable data distributions. This implies LoRA struggles to efficiently learn the additional complexity introduced by novel language data compared to simple domain adaptation, highlighting a boundary in its effectiveness. In addition to this, the additional complexity of reconstructing the target matrix usually implies longer training times as can be seen in Figure~\ref{fig:convergence-rate}. Therefore, a cost-benefit analysis is recommended: LoRA is superior for storage efficiency but may not be efficient if training speed is the primary constraint.

\subsubsection{Interactive Domain Adaptation in NMT}

Having established the effectiveness and boundaries of single-task LoRA adaptation, we decided to explore the possibility of interactively improving the performance of a given NMT model on a set of different domains.

To do this, we first pre-trained an NMT model on multiple datasets from different domains in order to obtain a generic translation model. Next, instead of simply training each new LoRA on a different domain to improve the performance of this generic model on that domain, we decided to pre-train a base LoRA on a wide variety of domains. The reason behind this approach was to obtain a generic LoRA that we could later use as a starting point for fine-tuning it on a variety of specialized domains, facilitating a smoother domain transition.

\begin{figure}[ht]
    \centering
    \begin{minipage}{0.5\textwidth}
        \centering
        \includegraphics[width=\textwidth]{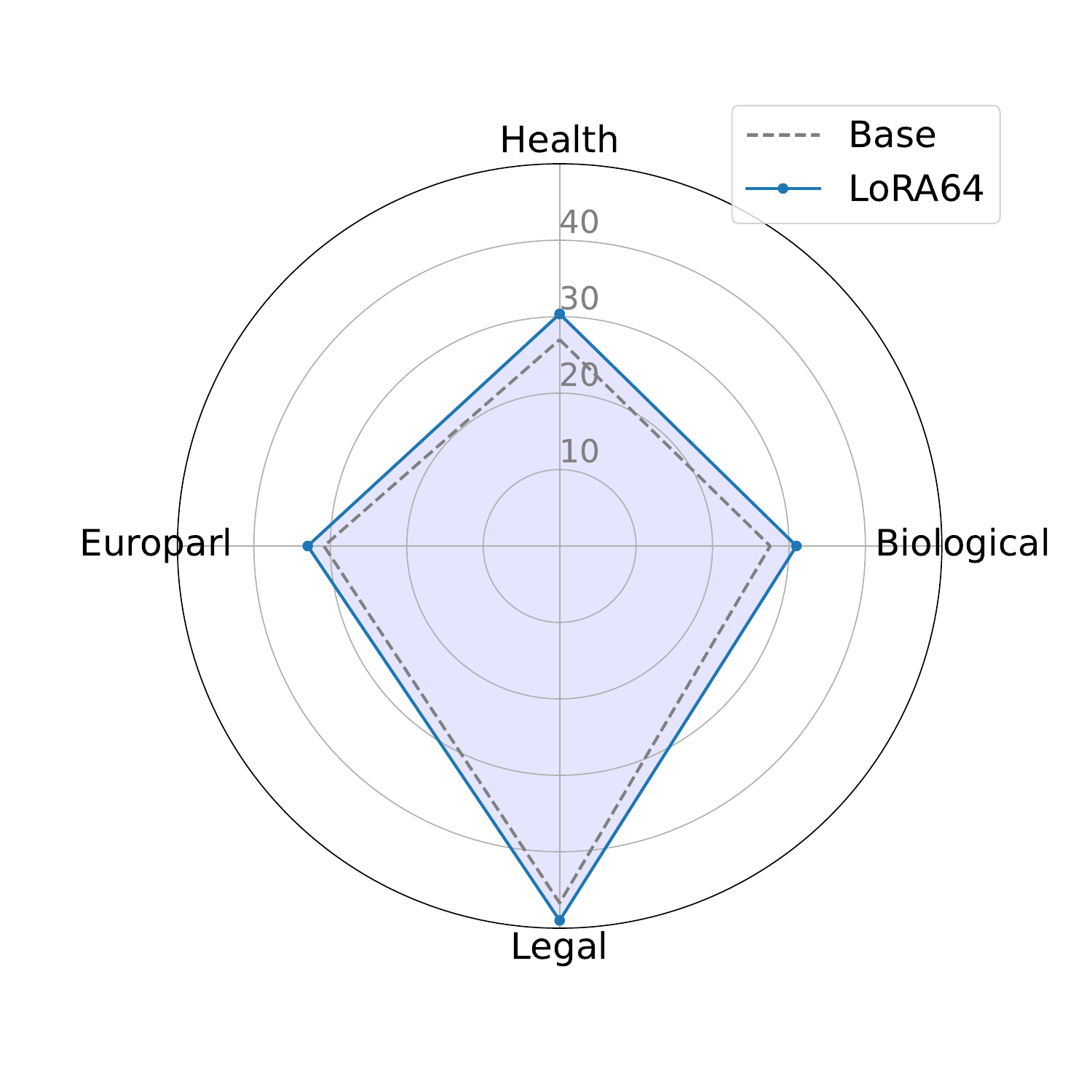}
        \label{fig:interactive-lora-rank64}
    \end{minipage}\hfill
    \begin{minipage}{0.5\textwidth}
        \centering
        \includegraphics[width=\textwidth]{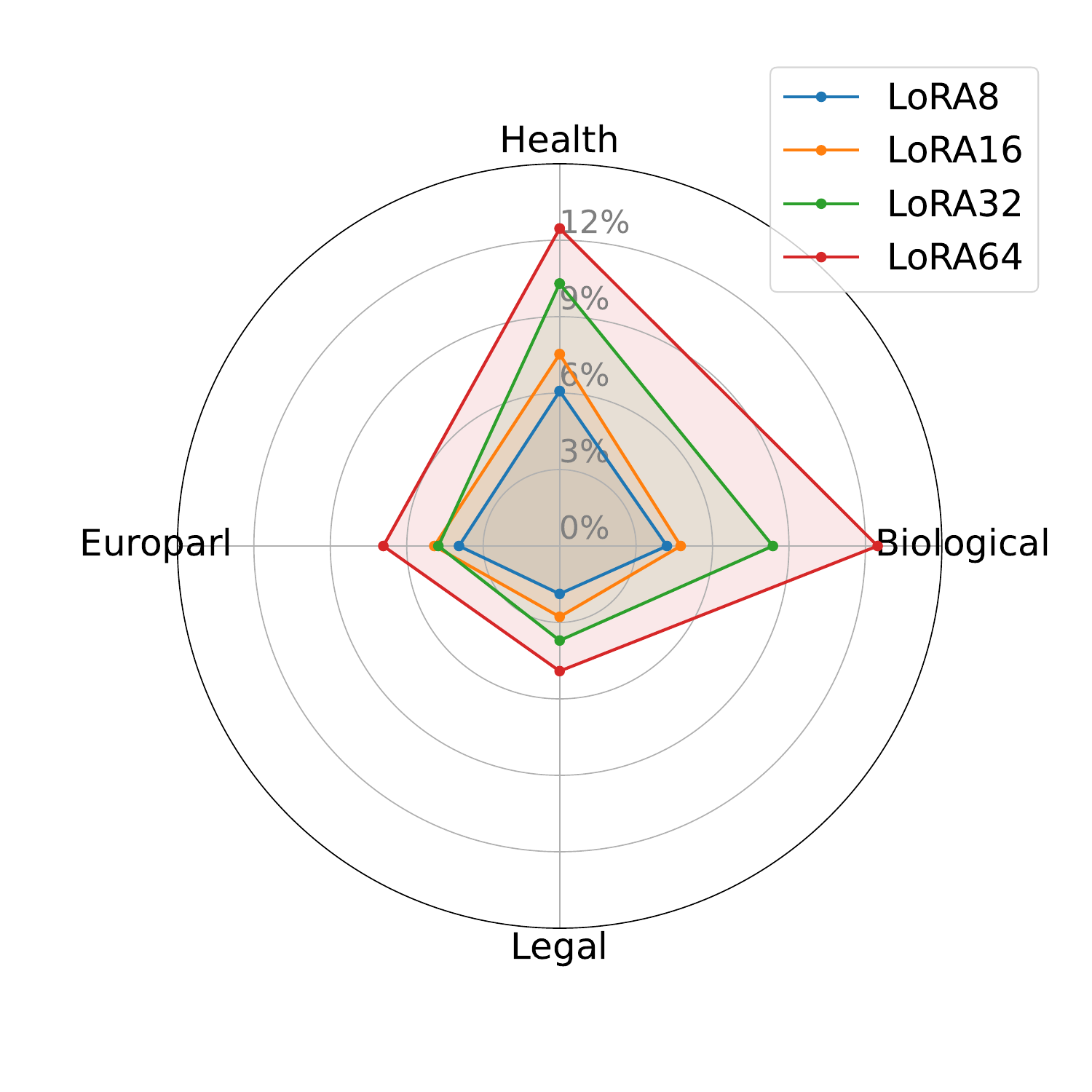}
        \label{fig:interactive-lora-rel}
    \end{minipage}
\caption{\textbf{Comparative analysis of interactive domains improvements:} The left figure contains the absolute BLEU scores for our base model (dashed line) as well as the area of performance improvement (blue area) achievable by varying the interpolation coefficients across four domains (health, biological, legal, and Europarl). The area represents the range of performance outcomes achievable by sweeping the scaling factor $\alpha$. The right figure details the relative performance gains (\%) after applying this LoRA strategy at different ranks (8, 16, 32, 64), showing a similar logarithmic trend in terms of performance and parameters as the previous experiments.}
\label{fig:interactive-lora}
\end{figure}

As it can be seen in Figure~\ref{fig:interactive-lora}, this strategy allowed us to use a linear combination of domain-specific LoRAs to interactively improve the performance, on the fly, on a given domain about the base model. 

In the left chart of Figure~\ref{fig:interactive-lora}, we observe the baseline performance of the base model (dashed line) alongside the performance-area per domain achieved after applying our LoRA strategy. Similarly, the right chart of Figure~\ref{fig:interactive-lora} extends this analysis by showing how the increase in parameter-resolution for each LoRA, from rank 8 to 64, significantly boosts the performance in each domain. Namely, this performance is incremented logarithmically about the number of parameters as seen in the previous section (See Section~\ref{subsec-lora-ft}). Therefore, the enhancements attributed to this LoRA approach also serve as a guide for the selection of an appropriate rank in finding a good balance between the performance gains and the expected computational efficiency.

Furthermore, we noticed that more often than not, this linear combination of domain-specific LoRAs acted as a sort of model ensemble that improved the performance across all domains, to the point of even surpassing the performance of some domain-specific LoRAs. However, it is also important to point out a few of the limitations that we encountered with this approach. The first limitation deals with the fact that some LoRAs contributed more strongly than others to a given domain, and despite scaling and normalizing each contribution per domain, this interaction between LoRAs seemed less and less linear and predictive as the distance between domains increased. Another limitation of this approach is that it does not attempt to overcome the effects of the catastrophic forgetting phenomenon so when we increase performance for a given set of domains, sometimes we end up worsening performance in other domains. This was especially noticeable when the domains were very different.

\subsubsection{Gradient-based regularization for low-rank matrices}

To deal with the aforementioned issues of forgetting during sequential adaptation, we now introduce our proposed gradient-based regularization strategy. The goal is to incorporate new knowledge into these low-rank matrices efficiently while mitigating catastrophic forgetting.

To achieve this, we designed a novel regularization strategy, as detailed in Section~\ref{sec-methodology}, which regularizes the low-rank matrices of the new model using the norm of the difference between the old and new weights, raised to a tunable power to provide finer control over the regularization process, and then, multiplying this term by the cumulative gradient of the previous task to re-weight the updates according to their importance for the previous task. 

\begin{figure}[ht]
    \centering
    \begin{minipage}{0.48\textwidth}
        \centering
        \includegraphics[width=\textwidth]{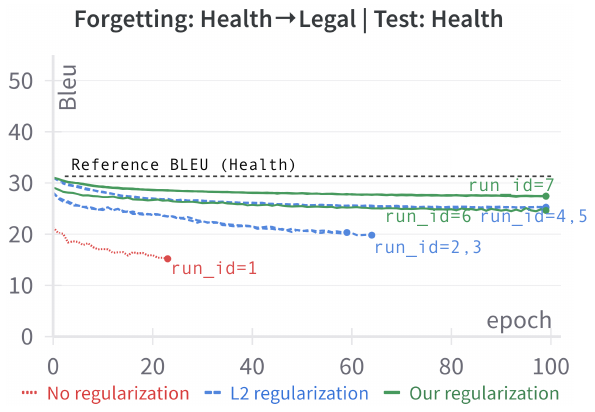}
        \label{fig:cf-domain-train}
    \end{minipage}\hfill
    \begin{minipage}{0.48\textwidth}
        \centering
        \includegraphics[width=\textwidth]{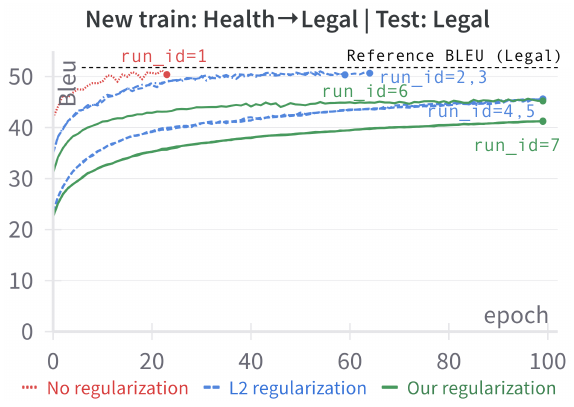}
        \label{fig:cf-domain-test}
    \end{minipage}
\caption{\textbf{CF regularization strategies in new domain scenarios:} The left figure shows the forgetting of the previous domain (health), while the right figure shows the learning of the new domain (legal). Our gradient-based regularization strategy (solid green line) was the strategy that was able to retain the past knowledge more consistently of all of them, despite losing a few performance points while learning the new domain, when compared to the no-regularization strategy. Then, the L2 regularization strategy acts as a middle ground between our approach and the no-regularized version. We conducted multiple runs for our method to analyze the sensitivity to the regularization coefficient $\gamma$. (\textit{run\_id}'s values are used to match training runs between both plots) }
\label{fig:cf-domains-grad}
\end{figure}

\begin{figure}[ht]
    \centering
    \begin{minipage}{0.48\textwidth}
        \centering
        \includegraphics[width=\textwidth]{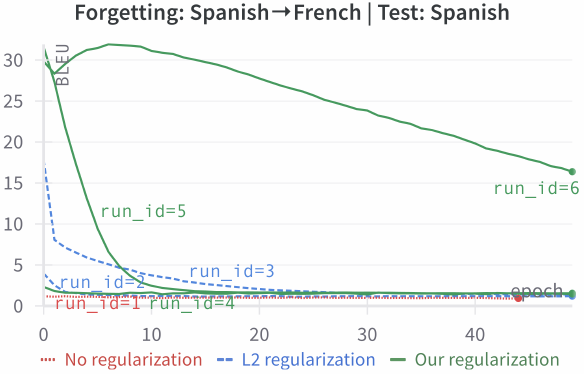}
        \label{fig:cf-language-train}
    \end{minipage}\hfill
    \begin{minipage}{0.48\textwidth}
        \centering
        \includegraphics[width=\textwidth]{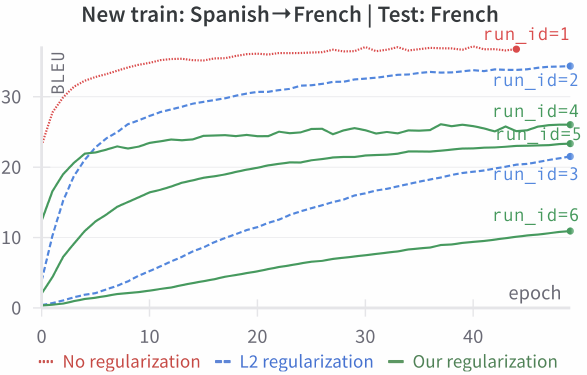}
        \label{fig:cf-language-test}
    \end{minipage}
\caption{\textbf{CF regularization strategies in new language-pair scenarios:} The left figure shows the forgetting of the previous language-pair (English-Spanish), while the right figure shows the learning of the new language-pair (English-French). Here, although no approach could be considered a success due to the complexity of the task, our gradient-based regularization (solid green line) proved to be the only one able to retain enough past knowledge while learning the new task (See \textit{run\_id=6}), a thing which we couldn't achieve with the other approaches due to their instant collapse of the previous knowledge during the very first iterations of training.}
\label{fig:cf-language-grad}
\end{figure}

Accordingly, we studied this approach from two different scenarios: The first scenario consisted of mitigating the catastrophic forgetting problem upon learning a new domain (health $\rightarrow$ legal) while keeping the language-pair fixed (See Figure~\ref{fig:cf-domains-grad}). And for the second and more challenging scenario, we attempted to add a new language-pair (English-French) without forgetting the original language-pair (English-Spanish), again, without having any past data available during training time (See Figure~\ref{fig:cf-language-grad}).

As a result, in Figure~\ref{fig:cf-domains-grad} we can see the performance of an NMT model being evaluated for two different domains (new and old) during the learning of the new domain. We report BLEU scores on the validation set during training to better capture the learning dynamics. Since both validation and test sets were randomly sampled from the same distribution, results were consistent across both. To begin with, the base model was initially trained exclusively on the health domain, and then, we fine-tuned it on the legal domain (with no past data available), using our gradient-based regularization for low-rank matrices to mitigate the catastrophic forgetting problem, along with no-regularization and the L2 regularization to serve as a comparison.

As it can be seen in Figure~\ref{fig:cf-domains-grad}, left (forgetting) and right (learning), there is a clear trade-off between the ability to learn a new domain and the resistance to forget known information. Generally speaking, our strategy (solid green line) can be quite resilient to the catastrophic forgetting effects during the learning of the new legal domain (from the health domain). However, it also exhibits an additional difficulty in learning the new task due to the added complexity of having to reconfigure the existing parameters from the previous task while learning this new task. Similarly, the L2 regularization (dashed blue line) shows less control over catastrophic forgetting in favor of a more efficient learning on the new task, establishing itself as an intermediate solution between our strategy and the non-regularized one (dashed red line), which as expected, stands out for its maximum efficiency in learning the new task while showing a complete disregard in preserving the knowledge from the previous task. Nonetheless, given that our approach presents more tunable aspects by means of the aforementioned hyperparameters, we expect it to be able to improve its learning rate by tweaking them a bit, without impacting its abilities to mitigate the catastrophic forgetting phenomenon. 

Next, we decided to explore a more challenging scenario, where instead of adding new domains, we added new language-pairs to our low-rank matrices for efficient task-switching scenarios(See Figure~\ref{fig:cf-language-grad}). This time, we started from a pre-trained model on an English-Spanish dataset, with the goal of adding a second language-pair, English-French, via low-rank matrices as an extreme scenario.

This task proved to be particularly difficult for all strategies (See Figure~\ref{fig:cf-language-grad}), given the fact that none of the tested strategies was able to successfully incorporate a second language-pair from scratch via low-rank matrices while retaining most of the past information (previous language-pair). We argue that the difficulty in making these two language-pairs coexist into our low-rank matrices (on top of a monolingual NMT model) could be related to a network saturation problem, since by not having enough parameters in which to store the new and the previous language-pair, we could be creating an artificial limitation. However, it is worth pointing out that our strategy turned out to be the only one that didn't produce an instant collapse of the previous language-pair (English-Spanish) during the very first iterations of the learning the new language-pair (See right figure~\ref{fig:cf-language-grad}, \textit{run\_id=6}), and instead, both the forgetting and the learning curve, turned out to be quite smooth and with a steady slope. Nonetheless, this is a very challenging task that will be studied more deeply in future work.

From these results, we demonstrate that we can effectively incorporate new knowledge into these low-rank matrices for an efficient task-switching strategy. However, there is no free lunch, since when using restrictions, we were not able to make the performance on new tasks (i.e., language or domain) as good as when no other restrictions were used during its learning. Finally, it is worth pointing out that, as expected, it seems to be significantly easier to incorporate new knowledge into these low-rank matrices when we are dealing with different domains or styles than when we are trying to add new language-pairs completely from scratch due to the additional complexity.

\section{Conclusion and Future Work}\label{sec-conclusion}

In this paper, we have introduced a holistic approach to address the continual learning problem in Neural Machine Translation (NMT), focusing on the use of Low-Rank Adaptation (LoRA) strategies to provide an efficient and effective solution to this problem. Our work contributes to the field of NMT by introducing a new paradigm that combines the strengths of low-rank matrices, interactive domain adaptation, and a gradient-based regularization approach to mitigate the catastrophic forgetting problem in this context.

Our research highlights the efficiency of task-switching strategies for NMT models using low-rank adaptation methods. This approach allows models to switch languages and domains highly efficiently, using far fewer parameters than traditional full-parameter fine-tuning strategies, which not only maximizes the computational resources but also maintains a performance-level comparable to more resource-intensive methods.

In addition to this, we have used a sort of Mixture of LoRA Experts in the form of a calibrated linear combination of domain-specific low-rank matrices, with which to make dynamic adjustments to the model's domain, style, and knowledge expertise, using a minimal amount of parameters.

Finally, we have introduced a gradient-based regularization strategy, specifically designed for this low-rank approach to mitigate the effects of the catastrophic forgetting phenomenon while new information is being incorporated into these low-rank matrices. 

However, our approach does come with its limitations. The balance between the rank of the matrices and the model's efficiency remains a critical consideration, as higher ranks, while improving the performance of our models, also reduce their parameter efficiency. Similarly, the calibration of multiple low-rank matrices requires careful consideration to avoid single-domain dominance and problems related to the catastrophic forgetting effects, and while our regularization strategy is effective in mitigating some of these problems, it also compromises performance on new tasks to some extent. We also note that our experiments relied on BLEU scores for broad comparability; future work should incorporate manual evaluation to better assess fluency and adequacy in fine-grained scenarios.

In summary, our work explores a new paradigm in the field of NMT for efficient, continual, and interactive machine translation, based on leveraging the efficiency of low-rank adaptation methods, to perform efficient task-switching strategies, and a gradient-based regularization method to incorporate new knowledge into our models without forgetting previous knowledge caused by the stability-plasticity dilemma.

\acks{Work supported by the Horizon 2020 - European Commission (H2020) under the SELENE project (grant agreement no 871467), the project LLEER, and the project Deep Learning for adaptive and multimodal interaction in pattern recognition (DeepPattern) (grant agreement PROMETEO/2019/121). We gratefully acknowledge the support of NVIDIA Corporation with the donation of a GPU used for part of this research.}

\vskip 0.2in
\bibliographystyle{theapa}
\bibliography{sample}

\end{document}